\documentclass{article}

\usepackage{comment}
\usepackage[final,nonatbib]{neurips_data_2023}
\usepackage{multirow}
\usepackage{wrapfig}
\usepackage[utf8]{inputenc} 
\usepackage[T1]{fontenc}    
\usepackage{hyperref}       
\usepackage{url}            
\usepackage{booktabs}       
\usepackage{amsfonts}       
\usepackage{nicefrac}       
\usepackage{microtype}      

\usepackage{ascmac}
\usepackage{graphicx}
\usepackage{makecell}
\usepackage[subrefformat=parens]{subcaption}
\usepackage{xcolor}

\usepackage{mdframed}
\usepackage{adjustbox}

\usepackage{pifont}
\title{CityRefer: Geography-aware 3D Visual Grounding Dataset on 
City-scale Point Cloud Data}

\author{%
\makecell{
Taiki Miyanishi$^{1,3}$\thanks{equal contribution}~~,\hspace{3pt}
Fumiya Kitamori$^{2*}$, \hspace{3pt}
Shuhei Kurita$^{3}$,\\
Jungdae Lee$^{2}$, \hspace{3pt}
Motoaki Kawanabe$^{1}$, \hspace{3pt}
Nakamasa Inoue$^{2}$}\\
\\
$^{1}$ATR, \hspace{3pt}
$^{2}$Tokyo Institute of Technology, \hspace{3pt}
$^{3}$RIKEN AIP\\
}

\begin{document}

\maketitle

\setcounter{footnote}{0}
\begin{abstract}
City-scale 3D point cloud is a promising way to express detailed and complicated outdoor structures.
It encompasses both the appearance and geometry features of segmented city components, including cars, streets, and buildings, that can be utilized for attractive applications such as user-interactive navigation of autonomous vehicles and drones.
However, compared to the extensive text annotations available for images and indoor scenes, the scarcity of text annotations for outdoor scenes poses a significant challenge for achieving these applications.
To tackle this problem, we introduce the {\it CityRefer dataset}\footnote{\url{https://github.com/ATR-DBI/CityRefer}} for city-level visual grounding. The dataset consists of 35k natural language descriptions of 3D objects appearing 
in SensatUrban~\cite{hu2022SensatUrban} city scenes 
and 5k landmarks labels synchronizing with OpenStreetMap.
To ensure the quality and accuracy of the dataset, all descriptions and labels in the CityRefer dataset are manually verified.
We also have developed a baseline system that can learn encoded language descriptions, 3D object instances, and geographical information about the city's landmarks to perform visual grounding on the CityRefer dataset.
To the best of our knowledge, the CityRefer dataset is the largest city-level visual grounding dataset for localizing specific 3D objects.
\end{abstract}


\section{Introduction}
\label{sec:intro}
Advancements in urban 3D scanning technologies, such as unmanned aerial vehicle photogrammetry and mobile laser scanning, enable the creation of accurate and photorealistic large-scale 3D scene datasets. Examples of such datasets include street-level datasets acquired by automobiles~\cite{behley2019semantickitti,Semantic3D,ParisLille3D,serna2014paris,Toronto3D,vallet2015terramobilita} and city-level datasets acquired by aerial vehicles~\cite{hu2022SensatUrban,li2020campus3d,rottensteiner2012isprs,varney2020dales,ye2020lasdu,zolanvari2019dublincity}.
However, while city-level photorealistic 3D scans have become practical and applicable in various fields like autonomous driving and unmanned vehicle delivery, the technology to comprehend city scenes through human-interactive linguistic representations is still in its early stages of development.
The ability to ground linguistic expressions to urban components is highly desired for interactive and interpretable applications, such as 
language-guided
autonomous driving and aerial drone navigation.
Achieving this requires the development of a 3D visual grounding dataset based on the city-scale point clouds, which presents a significant challenge.

3D Visual grounding is a 3D and language task that involves localizing objects in 3D scenes based on textual referred expressions. Compared to its 2D visual grounding counterparts~\cite{referitgame,refcocog,flickr30k,refcoco}, 3D visual grounding poses additional challenges. The expressions used in 3D visual grounding often require more information to localize object instances due to the rich context of 3D scenes, making the problem further complex. Recent studies have made remarkable progress in 3D visual grounding, focusing on determining the precise position of 3D objects given natural language descriptions~\cite{cai20223djcg,chen2021d3net,he2021transrefer3d,huang2021text,huang2022multi,jain2022bottom,roh2022languagerefer,yuan2022toward,yuan2021instancerefer,zhao20213dvg}.
However, many of these methods have been evaluated on 3D indoor datasets, which typically consist of point clouds of room scenes and labels for household objects~\cite{achlioptas2020referit_3d,chen2020scanrefer,scannetdata}.
Subsequently, Kolmet {\it et al.} \cite{Kolmet_2022_CVPR} proposed a  district-based visual grounding dataset on KITTI-360 \cite{liao2022kitti}. Nevertheless, the availability of 3D visual grounding datasets is still limited, especially in the context of aerial city-level 3D point clouds. 
Therefore, we aim to address this gap by creating a publicly available 3D visual grounding dataset based on aerial city-level 3D point clouds.

\begin{figure}[t]
\begin{center}
\includegraphics[width=14.05cm]{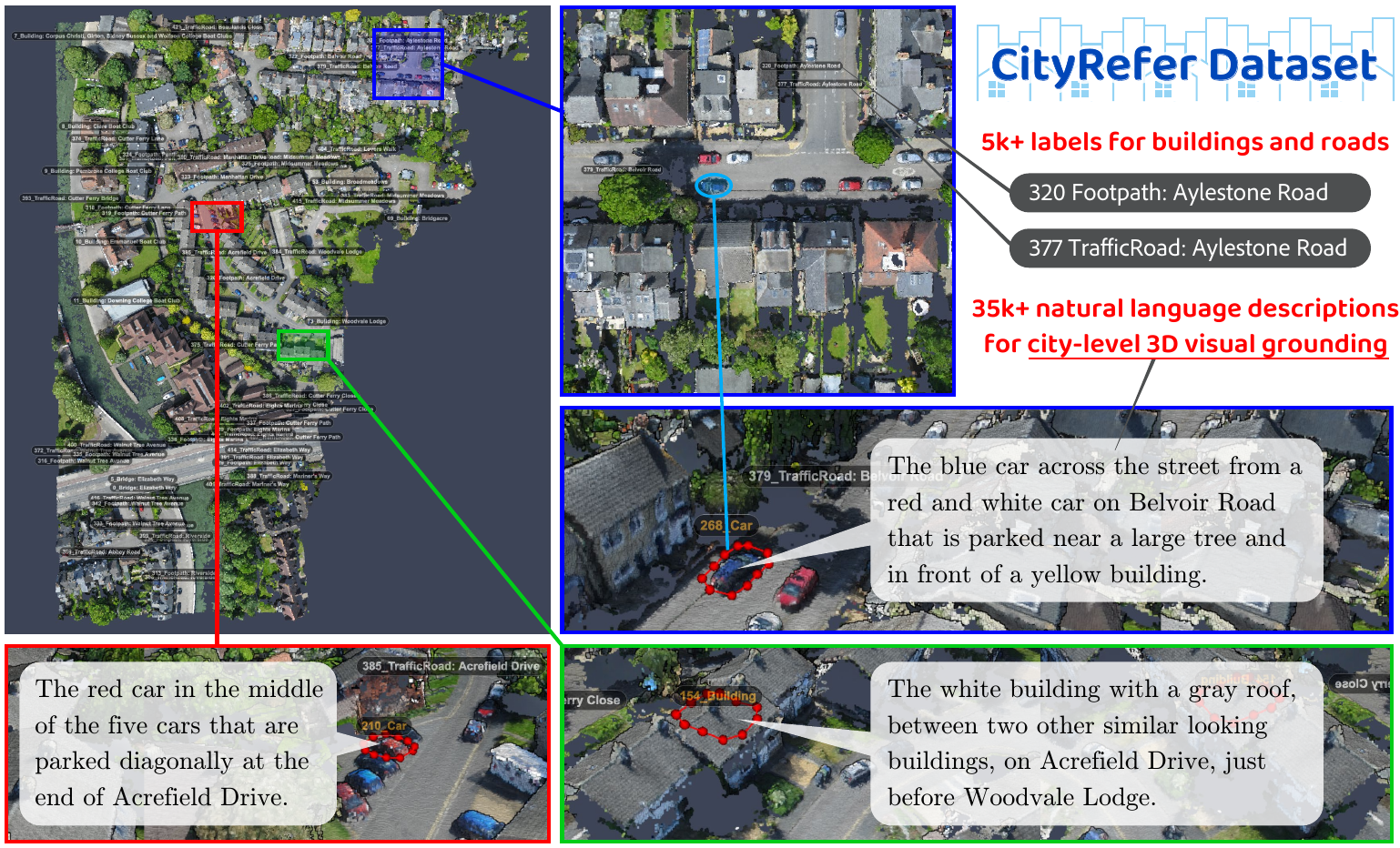}
\captionof{figure}{\textbf{The CityRefer dataset for city-level 3D visual grounding.}}
\label{fig:teaser}
\end{center}
\end{figure}

In this paper, we introduce the {\it CityRefer dataset} for city-level 3D visual grounding. Specifically, we provide 35k natural language descriptions to localize 3D objects in the SensatUrban~\cite{hu2022SensatUrban} environment as well as 5k labels of objects such as buildings and roads. Three example descriptions of a sample scene are shown in Figure~\ref{fig:teaser}.
As seen from the figure, city-level visual grounding is very challenging because a system is required to find objects from a wide city area while understanding the description of the target object and the relationships between relevant objects.
Although we used crowdsourcing to scale up the annotation, we needed to thoroughly filter out incorrect annotations by hand to finalize the dataset and ensure the quality of the annotations. The main contributions of the CityRefer dataset are summarized as follows.
{
\setlength{\leftmargini}{20pt}
\begin{enumerate}
\item We provide instance-wise segmentation masks for 5k objects including 1.8k landmark objects with their name labels. Examples include {\it Kem River and Baker Street, Aylestone Road, and Belvoir Road}. These labels were obtained from the spatial joint between SensatUrban and OpenStreetMap using our semi-automatic system (Section~3.1).

\item We provide 35k natural language descriptions for city-level visual grounding. These descriptions are manually attached using our interactive annotation system (Section~3.2).

\item We provide a baseline system that performs city-level 3D visual grounding. Because it is nontrivial to adapt previous visual grounding methods for our city-level dataset, we propose a simple but effective method that narrows the search area to find the target object by using geographical information.
\end{enumerate}
}

\begin{table*}[t]
\begin{center}
    \caption{Comparison of 3D visual grounding datasets. $N_{\mathrm{desc}}$ : Number of natural language descriptions. $\bar{L}_{\mathrm{desc}}$ : Average description length.
    $N_{\mathrm{points}}$ : Number of 3D points.
    }
	\footnotesize\begin{tabular}{llccccccc} 
        \toprule
& Dataset &
\hspace{-10pt}\thead{Human\\annot.} & $N_{\mathrm{desc}}$ & $\bar{L}_{\mathrm{desc}}$ & Area & Environment (Format) & $N_{\mathrm{points}}$  \\
\midrule
\multirow{5}{*}{\rotatebox[origin=c]{90}{Indoor}}
& REVERIE~\cite{Qi_2020_CVPR} &\hspace{-10pt}Yes & 21,702 & 18.0 & Rooms & Matterport3D (RGB)~\cite{Matterport3D}   & -   \\ 
& SUN-Spot~\cite{Mauceri_2019_ICCV}   &\hspace{-10pt}Yes & 7,987 & 14.1 & Rooms & SUN RGB-D~\cite{Song_2015_CVPR}   & -  \\
& SUNRefer~\cite{Liu_2021_CVPR}  &\hspace{-10pt}Yes & 38,495 & 14.1 & Rooms & SUN RGB-D~\cite{Song_2015_CVPR}   & - \\
& Nr3D~\cite{achlioptas2020referit_3d} &\hspace{-10pt}Yes & 41,503 & 11.4& Rooms  & ScanNet (3D Scan)~\cite{scannetdata}   & 242M   \\
& ScanRefer~\cite{chen2020scanrefer} & \hspace{-10pt}Yes & 51,583 &  20.3& Rooms & ScanNet (3D Scan)~\cite{scannetdata}    & 242M  \\
\midrule
\multirow{3}{*}{\rotatebox[origin=c]{90}{Outdoor}} &
TouchDown~\cite{Chen_2019_CVPR}  &\hspace{-10pt} Yes & 25,575 & 29.7 & Roadside & Google Street View (RGB) & -  \\
& KITTI360Pose~\cite{Kolmet_2022_CVPR}&\hspace{-10pt}No & 43,381 & 7.6  & Roadside & KITTI-360 (3D Scan)~\cite{liao2022kitti} & 1,000M  \\
& CityRefer (Ours) &\hspace{-10pt}Yes & 35,196 & 26.3 & City center & SensatUrban (3D Scan)~\cite{hu2022SensatUrban} & 2,847M \\
\bottomrule
	\end{tabular}
	\vspace{-0.2cm}
    \label{table:datasets}
\end{center}
\end{table*}

\section{Related Work}
\label{sec:related}

\noindent{\textbf{Visual Grounding Datasets for 3D Spaces.}} Over the years, there has been significant research interest in 3D visual grounding as summarized in Table~\ref{table:datasets}.
We first introduce two types of 3D visual grounding datasets, each focusing on a different level of the scene: the indoor and outdoor roadside. The 3D visual grounding dataset is created by annotating indoor or outdoor 3D datasets with linguistic descriptions.

\noindent \textbf{(i) Indoor scene level.} With the increasing availability of indoor 3D datasets~\cite{stanfordarmeni2017joint,dehghan2021arkitscenes,Matterport3D,scannetdata,Wald2019RIO,ramakrishnan2021hm3d,Song_2015_CVPR,xiazamirhe2018gibsonenv}, several visual grounding datasets have been proposed to address the demand for 3D scene understanding. REVERIE~\cite{Qi_2020_CVPR} comprises 10,318 panorama images captured across 86 indoor scenes and a total of 4,140 target objects. This dataset also provides 21,702 language instructions with rich textual annotations for guiding virtual agents within indoor photorealistic scenes of Matterport3D~\cite{Matterport3D}. The SUN-Spot~\cite{Mauceri_2019_ICCV} and SUNRefer~\cite{Liu_2021_CVPR} datasets focus on object localization in single-view RGB-D images within indoor environments from the SUN RGB-D dataset~\cite{Song_2015_CVPR}. Both datasets provide detailed language annotations indicating the spatial extent and location of objects in the images including object bounding boxes. Specifically, SUNRefer contains 38,495 language annotations for 7,699 objects in indoor RGB-D images.
Nr3D~\cite{achlioptas2020referit_3d} and ScanRefer~\cite{chen2020scanrefer} are standard 3D visual grounding datasets that are built on top of ScanNet~\cite{scannetdata}, a real-world 3D scene dataset with extensive semantic annotations.
However, these datasets mainly focus on the indoor visual grounding task.

\noindent \textbf{(ii) Outdoor scene level.} In recent years, a number of richly annotated outdoor 3D datasets have been created by scanning cities using sensors installed in cars and drones~\cite{behley2019semantickitti,Semantic3D,hu2022SensatUrban,li2020campus3d,ParisLille3D,Toronto3D,varney2020dales,Xie_2016_CVPR,zolanvari2019dublincity}. While there have been efforts to annotate outdoor 3D datasets with language descriptions for 3D visual grounding, the availability of such datasets is still limited compared to indoor ones. The TouchDown dataset~\cite{Chen_2019_CVPR} is aimed at text-guided navigation and spatial reasoning using real-life visual observations. It contains 9,326 examples of instructions and spatial descriptions in the visual navigation environment drawn from Google Street View.
KITTI360Pose~\cite{Kolmet_2022_CVPR} is a text-based position localization dataset in an outdoor 3D environment from the KITTI360 dataset~\cite{Xie_2016_CVPR}, 
which provides nine static scenes obtained by LiDAR scans. 
It is notable that the linguistic descriptions of KITTI360Pose are automatically generated by a sentence template with position description query pairs. 
While both TouchDown and KITTI360 datasets are based on the vehicle perspective and hence limited to the semantics from roadsides, our dataset is based on SensatUrban~\cite{hu2022SensatUrban}, which covers 3D semantics of the board city areas that are generated from aerial images by drones.

\noindent{\textbf{Learning Visual Grounding of 3D Scenes.}} To facilitate a deeper understanding of 3D scenes through language, there have been many efforts to connect languages and the 3D visual world, including 3D dense captioning~\cite{chen2021scan2cap,ijcai2022p194,Yuan_2022_CVPR}, 3D change detection~\cite{Qiu_2023_WACV}, and 3D visual question answering~\cite{abdelreheem2022scanents,Azumascanqa,etesam20223dvqa,ma2022sqa3d,zhao2022towards}. Specifically, 3D visual grounding that aims to locate an object in 3D space in response to a natural language query is the fundamental task of the 3D and language field~\cite{achlioptas2020referit_3d,chen2020scanrefer,wu2022eda}. 
Several approaches in the field of visual grounding use pre-computed 3D object detection or instance segmentation results, utilizing the point cloud features extracted from the corresponding 3D bounding boxes and segments~\cite{huang2021text,yuan2021instancerefer}.
However, the challenge of object recognition arises due to the low resolution of 3D data resulting from the reconstruction process. 
To overcome this limitation, recent studies have proposed the integration of both 2D images and 3D data~\cite{yang2021sat,jain2022bottom}. By combining the rich spatial information provided by 3D data with the detailed appearance cues derived from 2D images, these hybrid approaches aim to enhance robustness in the context of visual grounding.
Furthermore, there have been studies proposing methods that integrate 3D visual captioning and grounding, where both models are learned simultaneously to achieve synergistic effects~\cite{cai20223djcg,chen2021d3net}. 
These current 3D visual grounding methods mainly rely on two widely used indoor 3D visual grounding datasets~\cite{achlioptas2020referit_3d,chen2020scanrefer}. 
Several studies proposed visual grounding on remote sensing data but were limited to 2D images~\cite{Sun_2022_MM,Zhan_2023_TGRS}.
For these reasons, the performance of 3D visual grounding on outdoor 3D datasets remains unexplored. 
One of the initial studies~\cite{Kolmet_2022_CVPR} attempted to identify regions within the 3D point cloud based on textual queries on city-level 3D datasets but were limited to artificial language descriptions and did not fully use the geographic information 3D map despite the sparsity of city-level 3D point cloud data. In contrast, our method uses geographic information from 3D maps to achieve accurate city-level 3D visual grounding.

\begin{figure*}
\centering
\includegraphics[width=14.05cm]{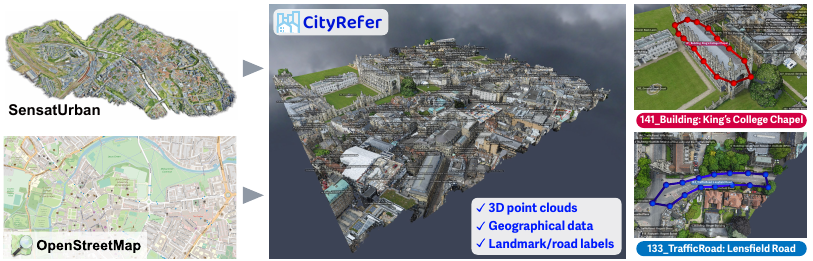}
\caption{\textbf{Stage 1 of dataset construction.} We perform a spatial join between the 3D SensatUrban environment and the 2D OpenStreetMap. The CityRefer dataset contains 1,850 landmark/road labels and geographical data. Examples include the landmark name {\it King's College Chapel} and the road name {\it Lansfield Road}.
}
\label{fig:stage_one}
\end{figure*}

\section{Dataset Construction}
\label{sec:dataset}
The CityRefer dataset consists of 1) instance-wise segmentation masks each with a label and geographical information, and 2) natural language descriptions for visual grounding. The 3D environment we use is the SensatUrban~\cite{hu2022SensatUrban}, which consists of photogrammetric point clouds of two UK cities covering 6 km$^2$ of the city landscape. Semantic segmentation masks are provided with the environment, and our study refines them to instance-level masks for visual grounding. The annotation proceeds in two stages: semi-automatic generation of instance-wise segmentation masks (Stage 1) and manual language annotation (Stage 2).

\subsection{Stage 1: Semi-automatic Generation of Instance-wise Segmentation Masks}
The goal of this stage is to generate segmentation masks for each 3D object as well as to attach labels of names with geographical data (longitude, latitude, and elevation) to each object. To achieve this, we perform a spatial join between the 3D environment and OpenStreetMap, from which we can obtain names and locations. Figure~\ref{fig:stage_one} shows an example result of this stage. As shown, regions of {\it King's College Chapel} and {\it Lensfield Road} are visualized precisely. The procedures are described in detail below.

\begin{figure*}
\centering
\includegraphics[width=12.5cm]
{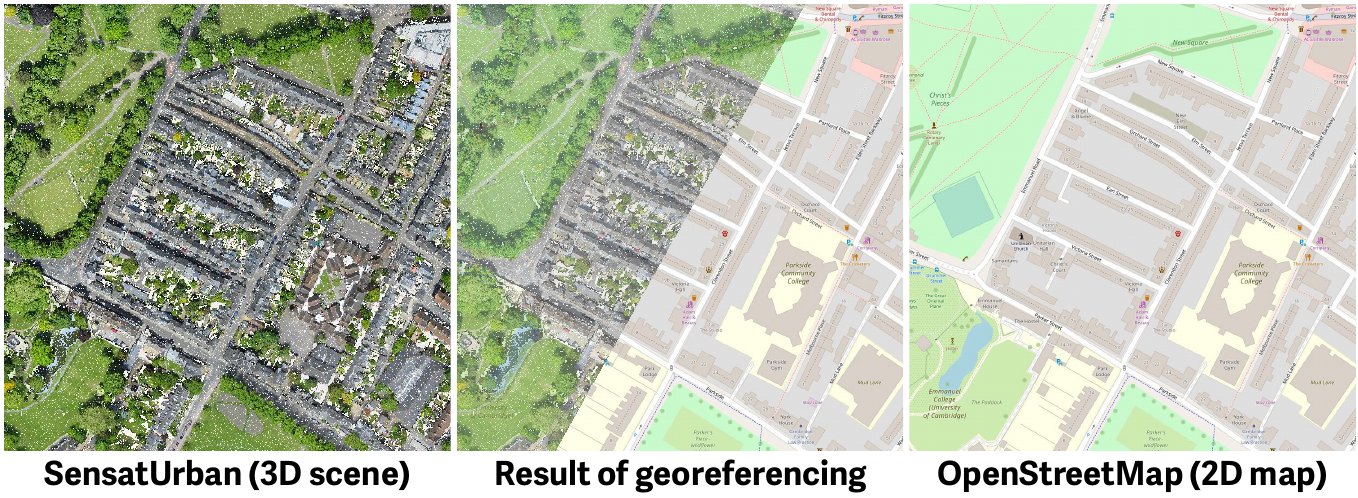}
\caption{Georeferencing of 3D environment and 2D map.}
\vspace{-10pt}
\label{fig:georeferencing}
\end{figure*}

\noindent \textbf{Georeferencing.}
The georeferencing is performed in the following three steps. 
First, given a block\footnote{We use 34 blocks (scenes) of Cambridge and Birmingham from SensatUrban.} of the 3D city, we create a top view image as shown in Figure~\ref{fig:georeferencing} (left), where the image size is fixed
to 2048 by 2048 pixels.
Second, we manually choose ten points of interest,
such as corner and landmark points, 
that are clearly visible in OpenStreetMap.
In this step, we also extract an image
of the 2D map as shown in Figure~\ref{fig:georeferencing} (right). 
Note that the coordinates of the ten points are manually annotated on both images.
Finally, we compute the geometric transformation between the two images
by using the transformation function from the scikit-image library.
We manually tune the hyper-parameters of transformation by visually verifying the results of the georeferencing. 
Here, the ten points of interest are also updated and tuned if needed. 
Figure~\ref{fig:georeferencing} (center) shows an example result in which the 3D scene and the 2D map are precisely joined.

\noindent \textbf{Generating instance-wise segmentation masks.} 
Given semantic segmentation masks with respect to 13 object categories, we refine them to instance-level masks.
We divide the categories into three groups as shown in Table~\ref{tab:group}.
For Group 1, we directly use geographic information obtained from OpenStreetMap to create filters for each instance. 
For example, for a segment of {\it Lansfield Road} in OpenStreetMap, 
we create a filter in the 3D scene based on 
the result of the georeferencing.
\def\hs{\hspace{-8pt}}
\begin{wraptable}[11]{hr}[5mm]{80mm}
\vspace{-10pt}
\caption{Three groups of categories.}
\vspace{-5pt}
\renewcommand{\arraystretch}{0.9}
\begin{tabular}{ccl}
\toprule
Grp.\hs& \hs Method\hs& Categories\\
\midrule
\hs\multirow{3}{*}{1}\hs&\hs\multirow{3}{*}{Filtering}\hs& Ground, HighVegetation,\\
& & Building, Bridge, Rail,\\
& & TrafficRoad, Footpath, Water \\
\midrule
\hs\multirow{2}{*}{2}\hs&\hs\multirow{2}{*}{Clustering}\hs& Wall, Parking,\\
& & StreetFurniture, Bike\\
\midrule
\hs3\hs&\hs{Detection}\hs& Car\\
\bottomrule
\end{tabular}
\label{tab:group}
\end{wraptable}
For Group 2, we apply the DBSCAN algorithm, a clustering method, to 3D points. It is not difficult to obtain accurate boundaries between instances by clustering because objects in this group are small and located separately. For Group 3, we used YOLOv7 to detect cars. The detection results are not perfect but are sufficient for creating the dataset. Note that we ask annotators to only use correctly segmented instances in the next stage. \vspace{-10pt}

\subsection{Stage 2: Manual Language Annotation}\label{sec:annotation}
\vspace{-0.2cm}
The goal of this stage is to collect natural language descriptions that describe the target object in a 3D scene for visual grounding. We prepare two interfaces: the language annotation interface and the quality control interface. The former is used to collect descriptions, and the latter is used to verify whether the collected descriptions are accurate. Below we present details of the interface implementation, the language annotation task, and the quality control procedure.

\begin{figure*}
\centering
\includegraphics[width=14.05cm]{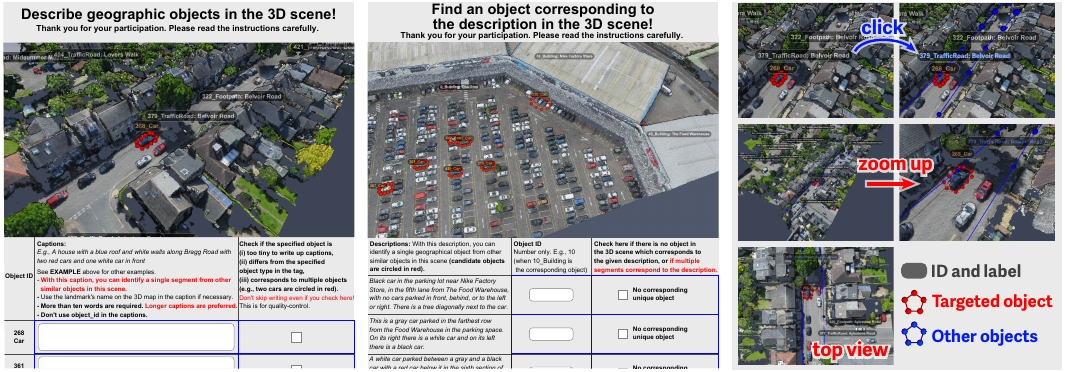}
\caption{\textbf{Stage 2 of dataset construction.} We ask annotators to describe the target object in the 3D scene so that the object can be uniquely identified based on the description. With our interface, annotators can see each object's region. Functions such as zooming and panning are also available.}
\label{fig:stage_two}
\end{figure*}

\noindent \textbf{Interface.} Figure~\ref{fig:stage_two} shows the interface we used for language annotation and verification. To interactively show 3D scenes to annotators, we implement the interface with Potree~\cite{schutz2016potree}, an open-source WebGL-based point cloud renderer that can process large-scale point cloud data. In the figure, the target object is highlighted in red. We provide interactive features such as zooming and panning as well as the labels of each object. The annotators can view the regions of each object by clicking the labels.
 
\textbf{Language annotation.}
We ask annotators to describe the target object with the following instructions.

\begin{mdframed}
You will see geographic objects in different 3D outdoor scenes. Please describe the objects in the 3D scene \textcolor{black}{\textbf{so that the objects can be uniquely identified} on the basis of your descriptions
and what you observed when submitting your responses.} Some of the information below (a
combination is preferred) should be included in your description:
{
\setlength{\leftmargini}{10pt}
\begin{itemize}
\vspace{-4pt}
\setlength{\itemsep}{3pt}
\setlength{\parskip}{0pt}
\setlength{\itemindent}{0pt}
\setlength{\labelsep}{5pt}
\item The object's appearance, {\it e.g., colors, shapes, materials.}
\item The object's location in the scene, {\it e.g., the parking lot is in front of Birmingham Library.}
\end{itemize}}
\end{mdframed}
\begin{mdframed}
{ \setlength{\leftmargini}{10pt} \begin{itemize}
\setlength{\itemsep}{3pt} \setlength{\parskip}{0pt} \setlength{\itemindent}{0pt} \setlength{\labelsep}{5pt}
\item The spatial relation between this object and other objects, {\it e.g., this building is the second one from the left.}
\end{itemize}
}
Imagine you and your friend live in a certain city, and you would like to ask your friend to find a geographic object in that city. Since there may be many similar geographical objects, the description of the object should be as unique as possible.
\end{mdframed}

To efficiently collect data, we show at most three target objects to annotators in a 3D scene. We also ask annotators to report segmentation errors and incorrect labels via another free-form text box and checkboxes.

\textbf{Quality control.}
To ensure the quality of the annotations, we ask another set of annotators to manually perform visual grounding with the following instructions.
\begin{mdframed}
\vspace{2pt}
You will see 1-3 descriptions for different geographic objects. Please choose the geographical object that best matches the description from the 3D scene (candidate objects are circled in red.) 
\end{mdframed}
\vspace{-2pt}
Based on the results, we remove incorrect descriptions. We measured the accuracy of the annotation with visual grounding again, and the correct response rate was 91.53\%. 

We used Amazon Mechanical Turk (MTurk) for annotation and quality control. 
There were 918 hours of work with a total cost of \$9,699 (the estimated hourly rate paid was \$10.56).
The total number of participating workers was 282.

\section{Dataset Statistics}
This section provides basic statistics of the CityRefer dataset in comparison to ScanRefer \cite{chen2020scanrefer} (an indoor dataset) and KITTI360Pose \cite{Kolmet_2022_CVPR} (a roadside dataset). 
We summarize the statistics in Table~\ref{tab:compare} and discuss them in detail below.

\begin{figure}[tbp]
\begin{minipage}[b]{0.66\columnwidth}
    \begin{center}
    \includegraphics[width=1.0\linewidth]{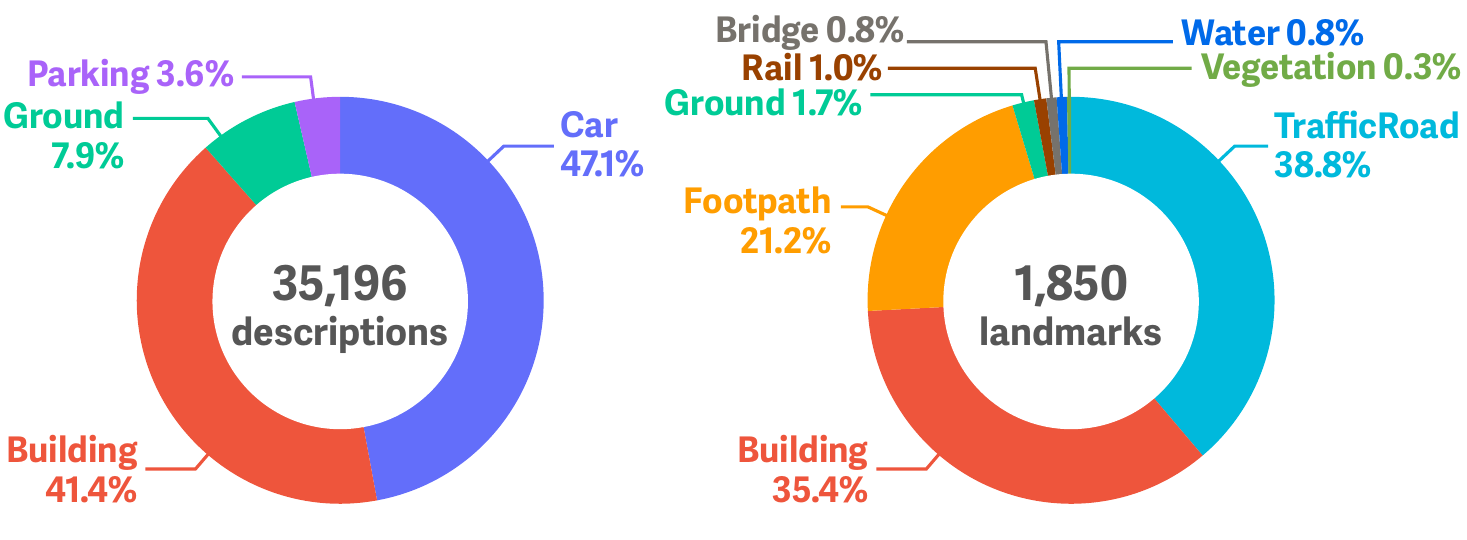}
    \caption{Object category distribution.}
    \label{fig:pie_chart}    
    \end{center}
\end{minipage}
\hspace{0.00\columnwidth}
\begin{minipage}[b]{0.33\columnwidth}
    \begin{center}
    \includegraphics[width=1.0\linewidth]{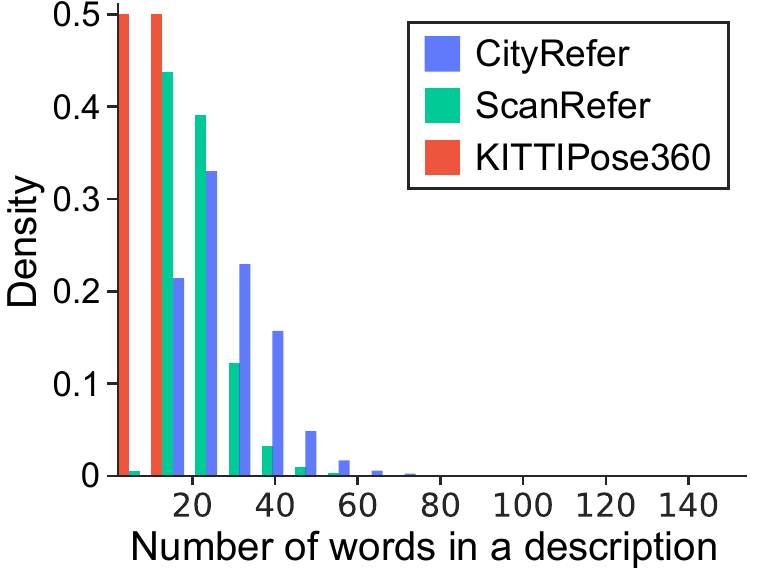}
    \caption{Description lengths}
    \label{fig:lengths}
    \end{center}
\end{minipage}
\end{figure}

\textbf{Target objects and descriptions.} 
The CityRefer dataset consists of 35,196 descriptions, each of which describes an object in the 3D scenes. 
The target objects fall under one of four categories: Car, Building, Ground, and Parking. 
The distribution is shown in Figure~\ref{fig:pie_chart} (left). 
Note that all of them are unnamed objects; that means that landmark objects such as famous buildings and roads, which can be identified by their names on OpenStreetMap, are excluded from the target objects. The distribution of description lengths is presented in Figure~\ref{fig:lengths}, along with those of ScanRefer and KITTI360Pose for comparison. Here, KITTI360Pose exhibits a sharp length distribution of the descriptions because they are automatically generated from a database with templates, e.g., {\it the pose is south of a gray road}. In contrast, our dataset provides moderate-length descriptions to perform visual grounding. To the best of our knowledge, the CityRefer dataset is the first large-scale dataset with manually annotated descriptions of city-level 3D scenes.

\textbf{Landmark objects.} There are 1,850 landmark objects with their names across seven categories: TrafficRoad, Building, Footpath, Ground, Rail, Bridge, Water, and Vegetation. The distribution is shown in Figure~\ref{fig:pie_chart} (right). Examples include {\it Senate House Hill}, {\it Wellhead Lane}, and {\it Parkside Police Station}. They are used to refer to target objects, {\it e.g., the gray rectangular building to the right of the parking lot next to St. John's College Chapel}.

%
%
\begin{table}[t]
\centering
\caption{\textbf{Comparison of datasets.} Area: Type of scanned area. Manual: Whether annotation is manual or not. Geo data: Availability of geographical data. $N_{\mathrm{desc}}$: Number of descriptions. $N_{\mathrm{obj}}$: Number of objects. $N_{\mathrm{landmark}}$: Number of landmark objects. $V$: Vocabulary size.
}
\vspace{2pt}
\begin{tabular}{l|ccccccc}
\toprule
Dataset & Area & Manual & Geo data & $N_{\mathrm{desc}}$ & $N_{\mathrm{obj}}$ & $N_{\mathrm{landmark}}$ & $V$\\
\midrule
ScanRefer \cite{chen2020scanrefer} & Indoor & $\checkmark$ & & 51,583 & 11,046 & 0 & 4,197\\
KITTI360Pose \cite{Kolmet_2022_CVPR} & Roadside & & & 43,381 & 6,800 & 0 & 41\\
CityRefer (Ours) & City center & $\checkmark$ & $\checkmark$ & 35,196 & 5,866 & 1,850 & 6,683\\
\bottomrule
\end{tabular}
\label{tab:compare}
\end{table}

%
%
\begin{figure}[t]
  \begin{subfigure}{0.245\textwidth}
    \begin{center}
    \includegraphics[scale=0.2]{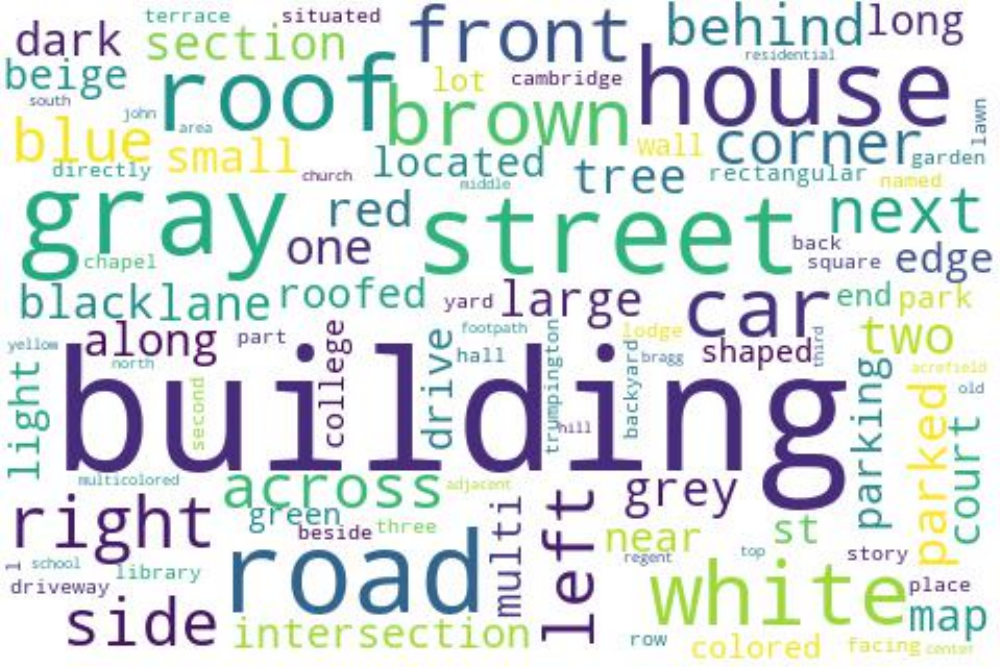}
    \caption{Building}
    \end{center}
  \end{subfigure}
  \begin{subfigure}{0.245\textwidth}
    \begin{center}
    \includegraphics[scale=0.2]{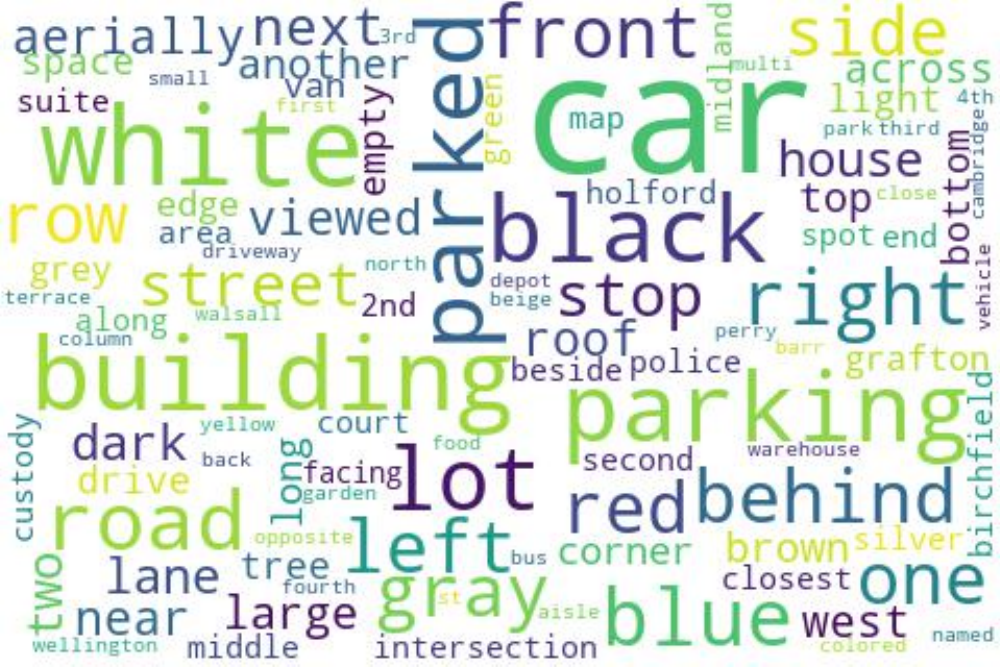}
    \caption{Car}
    \end{center}
  \end{subfigure}
  \begin{subfigure}{0.245\textwidth}
    \begin{center}
    \includegraphics[scale=0.2]{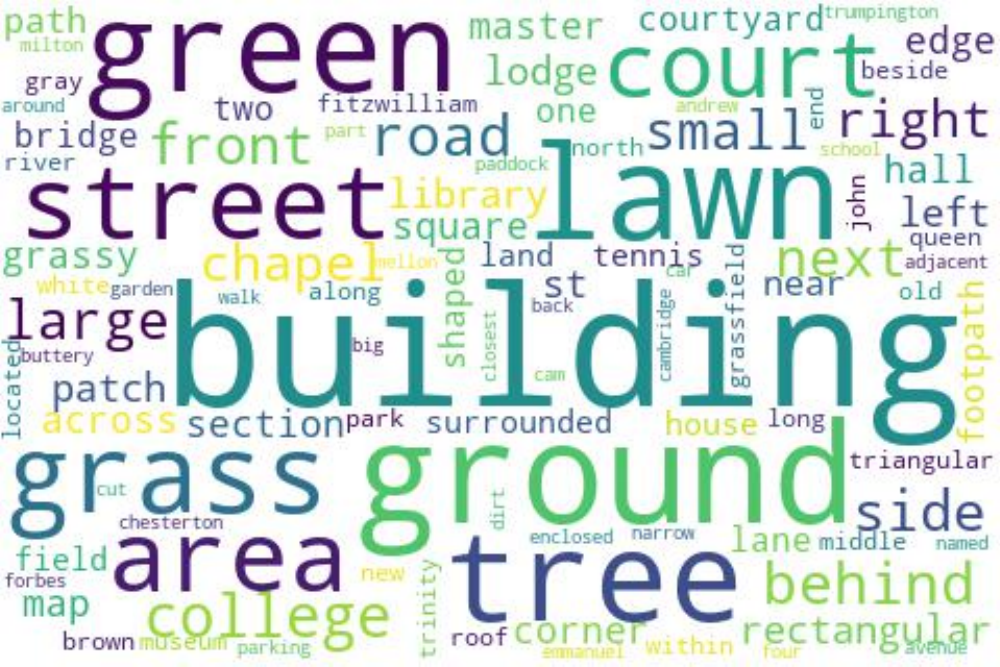}
    \caption{Ground}
    \end{center}
  \end{subfigure}
  \begin{subfigure}{0.245\textwidth}
    \begin{center}
    \includegraphics[scale=0.2]{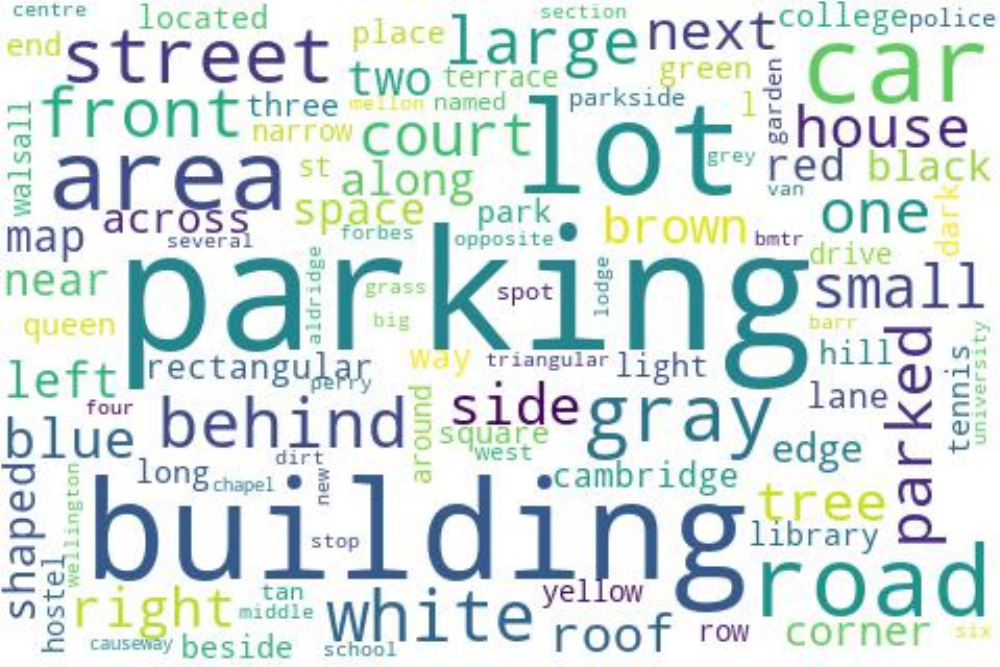}
    \caption{Parking}
    \end{center}
  \end{subfigure}  
  \caption{Visualization of word distributions.}
  \label{fig:word_cloud}
\end{figure}

\textbf{Words.} The vocabulary size of the CityRefer dataset is 6,683. Figure~\ref{fig:word_cloud} shows the visualization of word distributions. We can observe that words that specify object locations in the following three categories are frequently used: 1) colors {\it e.g., red, blue, gray}, 2) relative positions {\it e.g., right, left, across}, and 3) nearby objects {\it tree, street, road}. Compared with the template-based position descriptions of KITTI360Pose, our free-form descriptions contain more natural descriptions. 

\section{Experiments}
\label{sec:experiments}
Finally, we conducted experiments on the CityRefer dataset, focusing on two tasks: instance segmentation and visual grounding. To ensure comprehensive evaluation, we divided the dataset into three subsets: training, validation, and testing.
The data split is summarized in Table~\ref{tab:datasplit}, providing the number of descriptions ($N_{\mathrm{desc}}$), objects ($N_{\mathrm{obj}}$), and landmark objects ($N_{\mathrm{lmark}}$).

\subsection{City-level instance segmentation}
This experiment focuses on evaluating the segmentation performance. The objective is to generate segmentation masks with corresponding category labels for each object using 3D point clouds as input.

\textbf{Method.} 
We provide a PyTorch-based implementation of a baseline method using the SoftGroup++ model~\cite{vu2022softgroup}. Due to the larger number of 3D points (2,847M points) compared to previous datasets for 3D instance segmentation, we randomly sample 2\% of the 3D points and feed them into the model. To provide transparency and reproducibility, we summarize the training details along with manually tuned hyperparameters in the Appendix.

\textbf{Evaluation metrics.} 
We use average precision (AP) and mean recall (mRec) as our primary metrics. In addition, we report the top-$N$ AP and recall, with $N$ set to 50 and 25. 
Specifically, we refer to these metrics as AP$_{50}$, AP$_{25}$, mRec$_{50}$, and mRec$_{25}$.

\textbf{Results.} Table~\ref{tab:results_segmentation} shows the resulting performance for the four target categories.
We see that cars are relatively easier to segment compared to the segmentation of other objects.
However, the overall segmentation performance remains modest.
This is because the CityRefer dataset involves many similar instances placed near each other.

\subsection{City-level visual grounding}
In this experiment, we assess the visual grounding performance, which involves locating the target object within the 3D point clouds based on a given natural language description. To evaluate the performance of both instance segmentation and visual grounding separately, we assume that ground-truth instance segmentation masks are provided. 
We provide ten candidate answers, including the correct answer, for each description.

\noindent \textbf{Method.} We modified the InstanceRefer model~\cite{yuan2021instancerefer} to enable city-level visual grounding.
The modified model follows a four-step process to perform visual grounding.
First, we extract object features from each candidate instance by applying a four-layer sparse convolution network~\cite{Graham_2018_CVPR} with an average pooling layer to input 3D points.
Second, we use a one-layer bi-directional GRU (BiGRU)~\cite{chung2014empirical} to extract language features from the given description.
Third, object features and language features are concatenated and fed into another BiGRU to obtain visual-language features that represent the relationship between the description and the 3D objects.
Finally, we compute scores using a two-layer MLP.
These scores indicate the likelihood of a candidate instance being the correct grounding for the given description. 
We use the cross-entropy loss for training.
For more comprehensive information regarding the model architecture and training process, please refer to the Appendix.

%
%
\begin{table*}[t]
\begin{minipage}[t]{.37\textwidth}
\centering
\caption{Dataset split for training, validation, and testing}
\label{tab:datasplit}
\small
\begin{tabular}{l|ccc}
\toprule
Subset\hspace{-3.5pt}&\hspace{-3.5pt}$N_{\mathrm{desc}}$\hspace{-3.5pt}&\hspace{-3.5pt}$N_{\mathrm{obj}}$\hspace{-3.5pt}& \hspace{-3.5pt}$N_{\mathrm{lmark}}$\hspace{-3.5pt}\\
\midrule
Train\hspace{-3.5pt}& 23,586 & {3,931} & {1,106}\\
Val\hspace{-3.5pt}& 5,934 & {989} & {243}\\
Test\hspace{-3.5pt}& 5,676 & {946} & {501}\\
\midrule
Total &\hspace{-3.5pt}35,196\hspace{-3.5pt}&\hspace{-3.5pt}5,866\hspace{-3.5pt}&\hspace{-3.5pt}1,850\hspace{-3.5pt}\\
\bottomrule
\end{tabular}
\label{tabcompare}
\end{minipage}
\begin{minipage}[t]{.1\textwidth}
\end{minipage}
\begin{minipage}[t]{.6\textwidth}
\centering
\caption{Instance segmentation performance.}
\small
\begin{tabular}{l|cccccc}
\toprule
Target & AP & AP$_{50}$ & AP$_{25}$ & \hspace{-3.5pt}mRec\hspace{-3.5pt}&\hspace{-5pt}mRec$_{50}$\hspace{-5pt}&\hspace{-5pt}mRec$_{25}$\hspace{-5pt}\\
\midrule
Ground & 19.9 & 39.8 & 52.9 &\hspace{-5pt}28.0\hspace{-5pt} &\hspace{-5pt}47.2 &\hspace{-5pt}60.6\hspace{-5pt}\\
Building & 3.7 & 12.2 & 24.4 &\hspace{-5pt}7.2\hspace{-5pt}&\hspace{-5pt}16.8 &\hspace{-5pt}26.5\hspace{-5pt}\\
Parking & 5.9 & 17.0 & 48.3 &\hspace{-5pt}15.1\hspace{-5pt}&\hspace{-5pt}30.8 &\hspace{-5pt}59.0\hspace{-5pt}\\
Car & 35.3 & 55.0 & 69.4 &\hspace{-5pt}42.2\hspace{-5pt}&\hspace{-5pt}58.9 &\hspace{-5pt}70.9\hspace{-5pt}\\
\midrule
Average & 16.2 & 31.0 & 48.7 &\hspace{-5pt}23.1\hspace{-5pt}&\hspace{-5pt}38.4 &\hspace{-5pt}54.2\hspace{-5pt}\\
\bottomrule
\end{tabular}
\label{tab:results_segmentation}
\end{minipage}
\end{table*}

\textbf{Evaluation metrics.} 
We assess the accuracy of our predictions by comparing their intersection over union (IoU) with the ground truth values. 
Specifically, we focus on positive predictions that exhibit a higher IoU with the ground truth instances than a certain threshold $k$.
We use the Acc@$k$IoU metric, which is commonly used in the field of indoor 3D visual grounding research~\cite{chen2020scanrefer}.
For our experiments, we set the threshold value $k$ for IoU to 0.25.

\begin{figure*}
    \centering
    \includegraphics[width=1.0\linewidth]{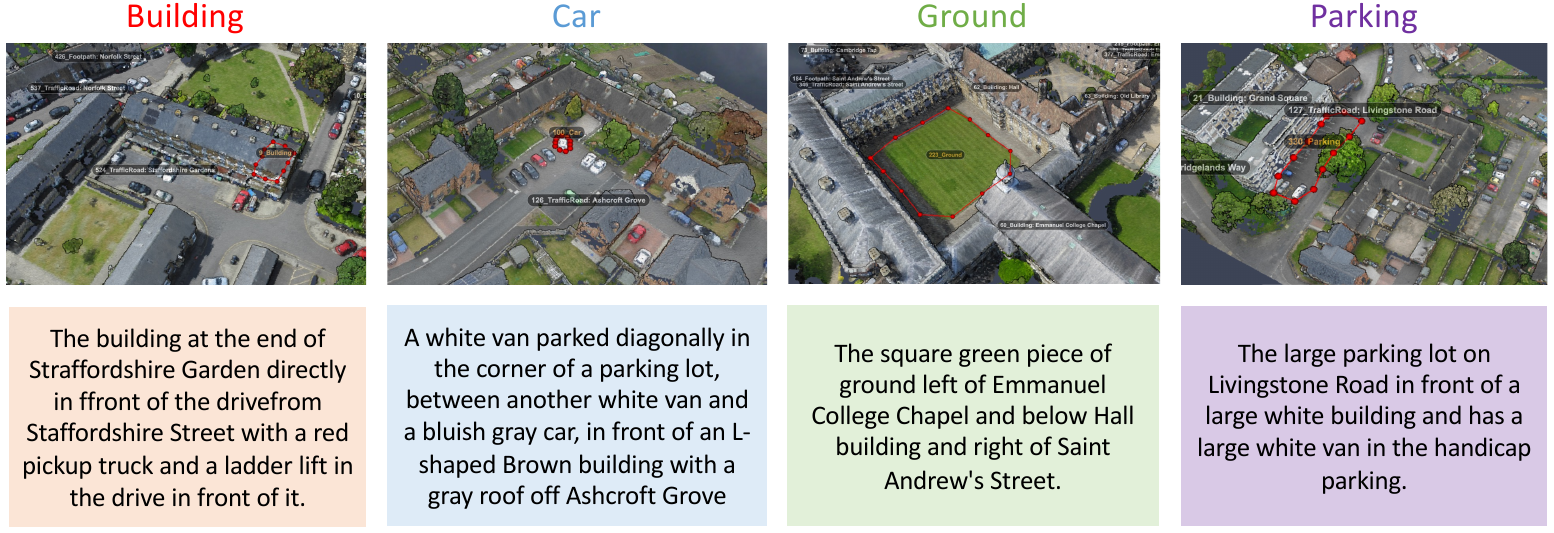}
    \caption{Qualitative examples.}
    \label{fig:qualitative}
\end{figure*}

\textbf{Results.}
Table~\ref{tab:results_3dvg} presents the visual grounding performance for the four target categories, comparing the results of Random (random guess) and Baseline (the method described above). 
Baseline + Land incorporates landmark features extracted from 3D points and the names of landmark objects, in addition to the object and language features.
We see that the baseline methods significantly perform better than the random guess but there is still a large gap between the system performance and {human performance (Acc. = 0.950)}.
This demonstrates that city-level visual grounding is a challenging task despite advances in learning technology. Developing large 3D-vision-language models for city-level visual grounding would be a potential future research direction.
Figure~\ref{fig:qualitative} illustrates qualitative examples of 3D visual grounding when incorporating landmark information. 
The examples demonstrate how the inclusion of landmark information aids in accurately identifying the target objects. 
This observation further reinforces incorporating landmark information is a promising way to improve the city-level visual grounding performance.

%
%
\begin{table*}[t]
\centering
\caption{City-level 3D visual grounding performance.}
\small
\begin{tabular}{l|cccc|c}
\toprule
Method  & Building &  Car & Ground & Parking & Overall \\
\midrule
Random & 0.103 $\pm$ 0.008 &  0.103 $\pm$ 0.006 & 0.091 $\pm$ 0.017 & 0.094 $\pm$ 0.010 & 0.101 $\pm$ 0.005 \\
Baseline  & \textbf{0.255} $\pm$ 0.005 & 0.282 $\pm$ 0.010 & 0.477 $\pm$ 0.024 & 0.835 $\pm$ 0.034 & 0.312 $\pm$ 0.006 \\
Baseline + Land  & \textbf{0.255} $\pm$ 0.008 & \textbf{0.298} $\pm$ 0.007 & \textbf{0.489} $\pm$ 0.009 & \textbf{0.853} $\pm$ 0.020 & \textbf{0.320} $\pm$ 0.005\\
\midrule
{Humans} & 0.947 & 0.956 & 0.937 & 0.945 & 0.950 \\
\bottomrule
\end{tabular}
\label{tab:results_3dvg}
\end{table*}

%
%

\section{Conclusion}
We have introduced the CityRefer dataset, a dataset for city-level visual grounding tasks. 
This dataset offers a comprehensive 3D environment where instance-wise segmentation masks, along with geographic data and labels, are provided. 
The creation of this environment involved performing a spatial joint between the SensatUrban environment and the OpenStreetMap, resulting in a rich and realistic urban setting.
We also provided 35k natural language descriptions to locate objects as well as baseline systems for instance segmentation and visual grounding.

\noindent \textbf{Limitations and future work.} In this work, we tackled visual grounding in the 3D environment of real cities. However, there are still gaps between the real world and the 3D environment. In particular, because all 3D scenes are static, the vocabulary of the CityRefer dataset is limited to words that specify static objects. 
To achieve more comprehensive and real-world applicability, it would be necessary to extend the dataset to include dynamic environments, where objects and scenes can change over time.
Furthermore, while our visual grounding system has demonstrated promising results, there is still room for improvement.
The main purpose of this work is to introduce a new dataset and indeed it is nontrivial to apply visual grounding systems for our dataset; however, developing more accurate systems would be worth pursuing in the future. We hope that the CityRefer dataset promotes further research and development as well as discussion on geography-aware learning technologies.

\textbf{Broader impacts.} Although the dataset is constructed on the basis of a publicly available 2D map (OpenStreetMap), visual grounding in general may result in privacy issues or racial and gender biases. The natural language descriptions we collect are carefully checked so that they do not include private information or offensive text.

\section*{Acknowledgements}
This work was supported by JST PRESTO JPMJPR22P8 and JPMJPR20C2, and by JSPS KAKENHI 22K12159.

\bibliographystyle{abbrv}
\bibliography{egbib}

\clearpage

\def\clearpagewithtitle#1{
\clearpage
\hrule height 4pt
\vskip 0.25in
\vskip -\parskip%
{
\centering
{\LARGE\bf
#1
\par}
}
\vskip 0.29in
\vskip -\parskip
\hrule height 1pt
\vskip 0.09in%
}

\clearpagewithtitle{CityRefer Datasheet}

\setcounter{page}{1}
\setcounter{section}{0}
\renewcommand{\thesection}{\Alph{section}}

We follow the guidelines of the datasheets for datasets~\cite{gebru2021datasheets}
to explain the composition, collection, recommended use case, and other details of the CityRefer dataset.

\section{Motivation}

\textbf{For what purpose was the dataset created?} \\
We created this CityRefer dataset to facilitate research toward city-scale 3D visual grounding.

\textbf{Who created the dataset (e.g., which team, research group) and on behalf of which entity (e.g., company, institution, organization)?} \\
This dataset was created by Taiki Miyanishi (ATR), Fumiya Kitamori (Tokyo Institute of Technology), Shuhei Kurita (RIKEN), Jungdae Lee (Tokyo Institute of Technology), Motoaki Kawanabe (ATR), and Nakamasa Inoue (Tokyo Institute of Technology).

\textbf{Who funded the creation of the dataset?} \\
This work was supported by JST PRESTO JPMJPR22P8 and JPMJPR20C2, and JSPS KAKENHI
22K12159.


\if[]
\section{Author Statement}
We bear all responsibilities for the licensing, distribution, and maintenance of our datasets.
\fi

\section{Composition}
\textbf{What do the instances that comprise the dataset represent?} \\
CityRefer contains descriptions for 3D visual grounding on large-scale point cloud data.
We do not provide the 3D point cloud data, which can be downloaded from the official site of SensatUrban~\cite{appendix_hu2022SensatUrban}.

\textbf{How many instances are there in total (of each type, if appropriate)?} \\
There are 5,866 objects on the 3D map with their instance masks.
There are 35,196 natural language descriptions for visual grounding.

\textbf{Does the dataset contain all possible instances or is it a sample (not necessarily random) of instances from a larger set?} \\
Landmark objects are sampled from OpenStreetMap\footnote{\url{https://www.openstreetmap.org}}.
They are representative of all the possible geographical objects.

\textbf{Is there a label or target associated with each instance?} \\
Yes.

\textbf{Is any information missing from individual instances?} \\
No.

\textbf{Are relationships between individual instances made explicit (e.g., users’ movie ratings, social network links)?} \\
Yes. We provide metadata for each object.

\textbf{Are there recommended data splits (e.g., training, development/validation, testing)?} \\
Yes. We provide metadata of data splits.

\textbf{Are there any errors, sources of noise, or redundancies in the dataset? } \\
Please refer to the ``Quality control'' in Sec.~\ref{sec:annotation}.

\textbf{Is the dataset self-contained, or does it link to or otherwise rely on external resources (e.g., websites, tweets, other datasets)?} \\
We follow prior work~\cite{appendix_chen2020scanrefer} and provide descriptions for 3D visual grounding.

\textbf{Does the dataset contain data that might be considered confidential?} \\
No.

\textbf{Does the dataset contain data that, if viewed directly, might be offensive, insulting, threatening, or might otherwise cause anxiety?} \\
No.

\section{Collection Process}
The collection procedure, preprocessing, and cleaning are explained in Sec.~\ref{sec:dataset} of our main paper.

\textbf{Who was involved in the data collection process (e.g., students, crowdworkers, contractors), and how were they compensated (e.g., how much were crowdworkers paid)?} \\
Data collection and filtering are done by crowdworkers. 
Data curation is done by coauthors.

\textbf{Over what timeframe was the data collected?} \\
The data was collected between January 2023 to April 2023.

\section{Uses}
\textbf{Has the dataset been used for any tasks already?} \\
Yes. We have used the CityRefer database for city-scale 3D visual grounding. 
Please refer to Sec.~\ref{sec:experiments} in our main paper.

\textbf{Is there a repository that links to any or all papers or systems that use the dataset?} \\
Yes.

\textbf{What (other) tasks could the dataset be used for?} \\
Our dataset is primarily intended to facilitate research in 3D visual grounding. However, it can also be broadly applicable to 3D and language tasks such as 3D object retrieval, 3D question answering, 3D dense captioning, language-guided navigation, embodied question answering, etc.

\textbf{Is there anything about the composition of the dataset or the way it was collected and preprocessed/cleaned/labeled that might impact future uses?} \\
Nothing.

\textbf{Are there tasks for which the dataset should not be used?} \\
It should not be used as a tool to monitor individuals without regard for their privacy.

\section{Distribution}
\textbf{Will the dataset be distributed to third parties outside of the entity (e.g., company, institution, organization) on behalf of which the dataset was created?} \\
Yes.

\textbf{How will the dataset will be distributed (e.g., tarball on website, API, GitHub)?}\\
The CityRefer dataset and our baseline code can be downloaded from our webpage\footnote{\url{https://github.com/ATR-DBI/CityRefer}} under CC-BY4.0 license and MIT license, respectively.

\textbf{Have any third parties imposed IP-based or other restrictions on the data associated with the instances?} \\
No.

\textbf{Do any export controls or other regulatory restrictions apply to the dataset or to individual instances?} \\
No.


\section{Maintenance}
\textbf{Who will be supporting/hosting/maintaining the dataset?} \\
The authors will be supporting, hosting, and maintaining the dataset.

\textbf{How can the owner/curator/manager of the dataset be contacted (e.g., email address)?} \\
The contact email address can be found on our website.

\textbf{Is there an erratum?} No. We will provide the erratum as soon as the need arises.

\textbf{Will the dataset be updated (e.g., to correct labeling errors, add new instances, delete instances)?} \\
Yes. 

\textbf{If the dataset relates to people, are there applicable limits on the retention of the data associated with the instances (e.g., were the individuals in question told that their data would be retained for a fixed period of time and then deleted)?} \\
N/A.

\textbf{Will older versions of the dataset continue to be supported/hosted/maintained?} \\
Yes.

\textbf{if others want to extend/augment/build on/contribute to the dataset, is there a mechanism for them to do so?}
N/A.




\clearpagewithtitle{CityRefer: Supplementary Material}
\setcounter{section}{0}

This is supplementary material for the paper: {\it CityRefer: Geography-aware 3D Visual Grounding Dataset on City-scale Point Cloud Data}.
We present additional details of the dataset, instance segmentation, and 3D visual grounding. We also describe additional ablation studies with qualitative results.

\section{Dataset details}
\subsection{Data Collection Interface}
We developed an annotation website on the Amazon Mechanical Turk platform for language annotation and manual 3D visual grounding. Figure~\ref{fig:cap_ann_inter} shows the annotation interface for describing target objects. To collect natural language descriptions about the target object in the 3D map, we ask annotators to describe the target object following the given instructions written in the annotation interface. We also provide examples of geographical objects and their corresponding descriptions.
For quality-control purposes, we also ask annotators to check if the specified object:
(i) is too tiny to write captions for,
(ii) differs from the specified object type in the tag, or
(iii) corresponds to multiple objects (e.g., two cars are red-lined).
Authors manually confirmed and removed the incorrect data for 5 out of 6 objects checked by annotators.

\subsection{Quality Control Details} 
\begin{figure*}
\centering
\includegraphics[width=12cm]{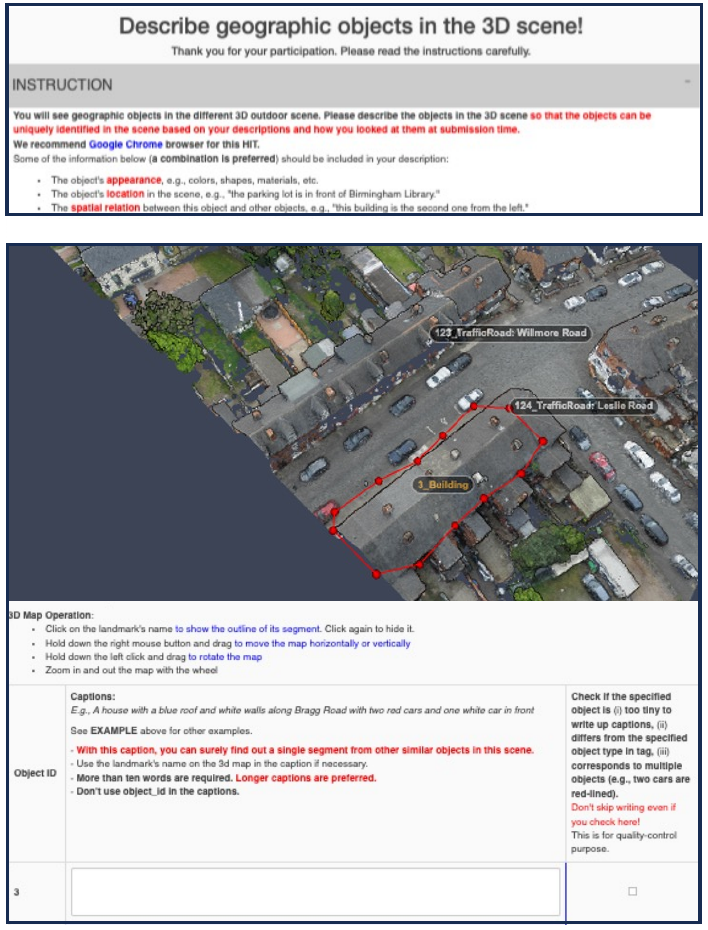}
\caption{Annotation interface for collecting object descriptions.
The top shows an instruction and annotation example part, and the bottom shows the part where workers input descriptions corresponding to given object IDs.
}
\label{fig:cap_ann_inter}
\end{figure*}

\begin{figure*}
\centering
\includegraphics[width=12cm]{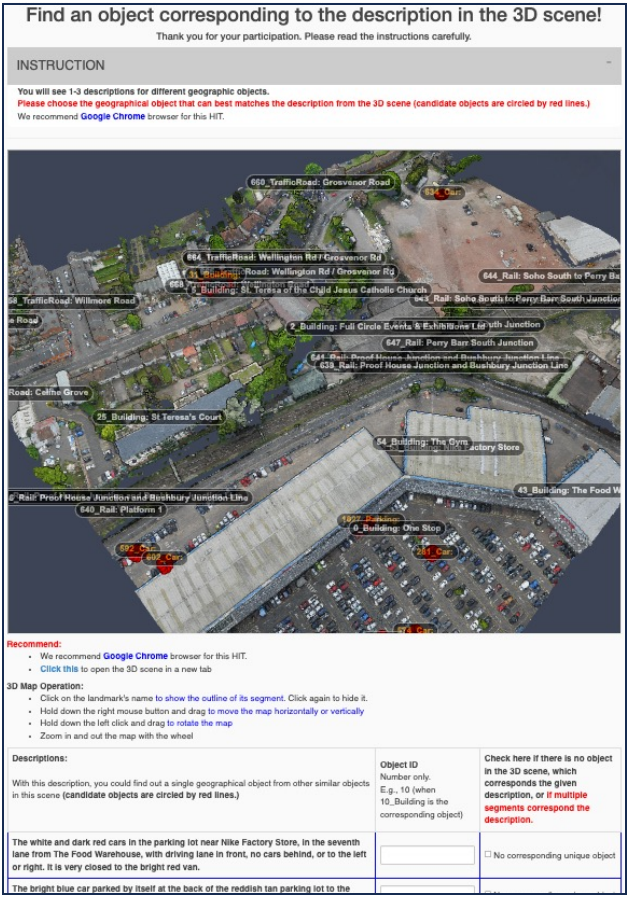}
\caption{Annotation interface for checking object's descriptions to perform manual 3D visual grounding. Workers input object ID in the 3D map, corresponding to given descriptions.}
\label{fig:vg_ann_inter}
\end{figure*}

To further improve the quality of the annotations, we filter out inappropriate descriptions using a manual 3D visual grounding website and re-annotate them. After collecting the initial descriptions, we present the 3D map, along with the corresponding object names and IDs, to the workers. Figure~\ref{fig:vg_ann_inter} shows the annotation interface for 3D visual grounding. The workers are instructed to enter the object IDs that best match the provided descriptions for the 3D map.
In addition, they are prompted to check a box if no object in the 3D map matches the description or if multiple objects correspond. We discard such incomplete descriptions and re-annotate the corresponding objects using the annotation website used during the initial annotation step, as shown in Figure~\ref{fig:cap_ann_inter}. To ensure comprehensive coverage, we collect six descriptions for each object, thereby capturing multiple perspectives and linguistic variations.

\section{Instance Segmentation Details}
\subsection{Architecture}
\begin{figure*}[ht]
    \centering\includegraphics[width=1.0\linewidth]{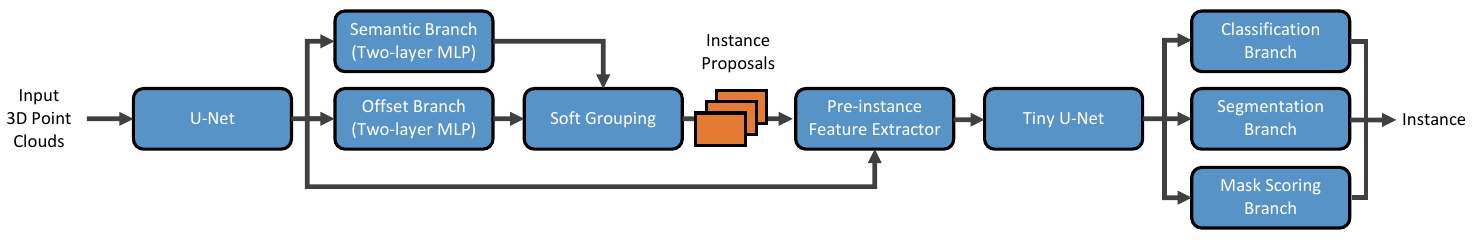}
    \caption{SoftGruop++ architecture \cite{appendix_vu2022softgroup} for instance segmentation.}
    \label{fig:softgroup}
\end{figure*}

We used the SoftGroup++ architecture~\cite{appendix_vu2022softgroup}, an extension of SoftGroup for our instance segmentation task.
Figure~\ref{fig:softgroup} shows the overview of SoftGroup++.
The approach consists of two main stages: bottom-up grouping and top-down refinement. Initially, point features are extracted from the input point clouds using a U-Net backbone. Next, semantic scores and offset vectors are predicted by the semantic and offset branches, respectively. 
A soft grouping module then uses these predictions to generate instance proposals. The feature extractor layer extracts backbone features from these proposals, which a tiny U-Net subsequently processes. 
Finally, the classification, segmentation, and mask-scoring branches are used to derive the final instances.
For our experiments, we used the official implementation\footnote{\url{https://github.com/thangvubk/SoftGroup/tree/softgroup++}} of SoftGroup++ and customized the dataset configuration to suit the SensatUrban dataset.

\subsection{Training}
We conducted training of SoftGroup++ on a large-scale point cloud dataset containing a total of 5,866 objects belonging to the `Ground,' `Building,' `Parking,' and `Car' categories. To achieve this, we adopted the configuration used, for instance segmentation in STPLS3D~\cite{appendix_Chen_2022_BMVC}, a large-scale synthetic 3D point cloud dataset. To ensure computational efficiency, we downsampled the input point clouds uniformly to a ratio of 1/50. Additionally, we cropped the point clouds into non-overlapping blocks, each covering an area of $250m^2$.
During training, we initialized the learning rate to 4e-3 and used a cosine annealing scheduler to adjust it. The training process was executed on a single node equipped with four V100 GPUs, using FP16 mixed precision for improved computational performance.
All hyperparameters used in our training setup are listed in Table~\ref{table:3dseg_hyperparameter}.

\begin{table*}[ht]
 \caption{Hyperparameters for training the 3D instance segmentation model.}
 \label{table:3dseg_hyperparameter}
 \centering
  \begin{tabular}{r|l}
   \toprule
    Hyperparameter & Value \\
    \midrule
    Training epoch & 108 \\
    Optimizer & Adam~\cite{appendix_kingma2014adam} \\
    Learning rate & 4e-3 \\ 
    Batch size per GPU & 4 \\
    Voxel size & 0.33 \\
    Number of semantic classes & 4 \\
   \bottomrule
  \end{tabular}
\end{table*}

\section{3D Visual Grounding Details}
We present details of our geography-aware 3D visual grounding model. 
All the neural networks in our implementation were developed using PyTorch v1.31. As our baseline method, we relied on the code provided by InstanceRefer~\cite{appendix_yuan2021instancerefer}\footnote{\url{https://github.com/CurryYuan/InstanceRefer}} and made appropriate modifications to suit the city-scale 3D visual grounding task.

\subsection{Architecture}
\begin{figure*}[t]
    \centering\includegraphics[scale=0.55]{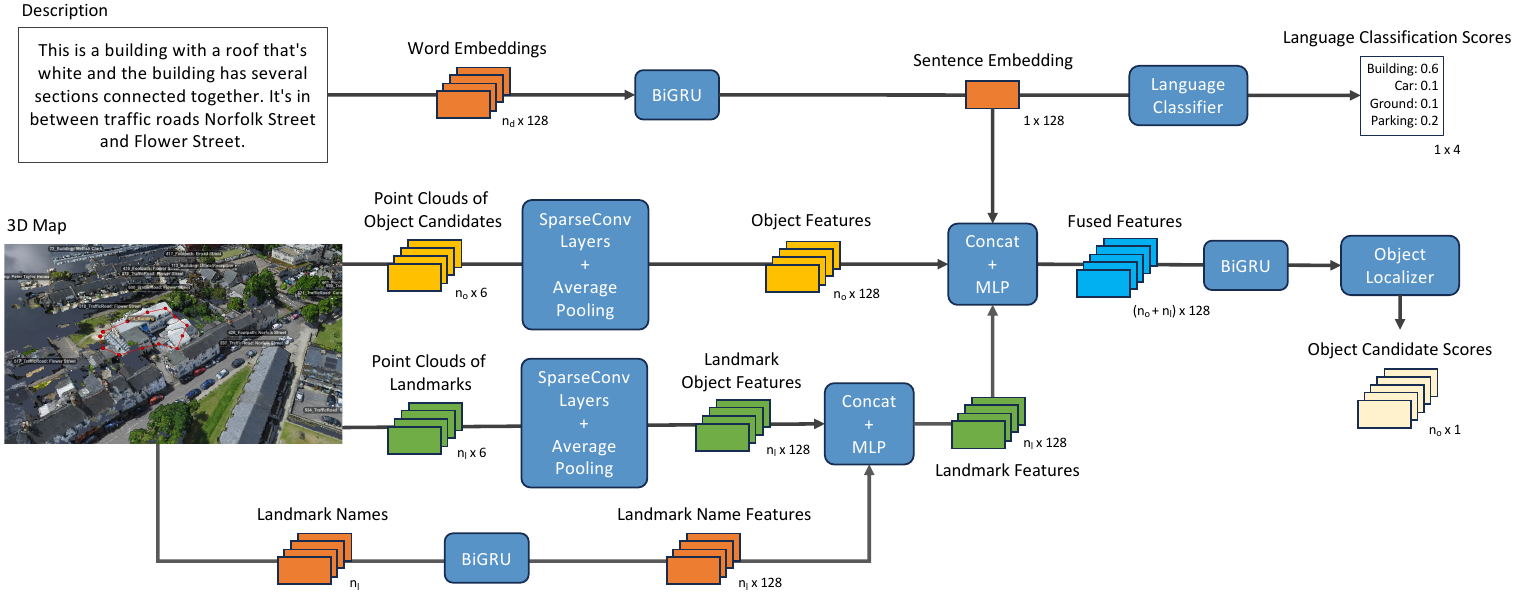}
    \caption{CityRefer architecture for 3D visual grounding.}
    \label{fig:3dvg_baseline}
\end{figure*}

We developed a CityRefer model consisting of language \& 3D object encoders along with an object localization module.
Figure~\ref{fig:3dvg_baseline} provides an overview of our model architecture. 
In formal terms, we define the inputs to our model as follows: language description $D$, landmarks $L$, and candidate objects $O$ in the 3D map.

\textbf{Language encoder.}
To process the description, we begin by tokenizing it into tokens $\{w_i\}_{i=1}^{n_d}$ using the BertTokenizer\footnote{\url{https://huggingface.co/docs/transformers/model_doc/bert}}.
We then perform projection to obtain word representations $W \in \mathcal{R}^{n_d \times 128}$. Here, $n_d$ represents the number of tokens in the description. These representations are subsequently fed into a one-layer bidirectional GRU (BiGRU) for word sequence modeling. We use the first hidden state from the BiGRU as the sentence embedding for the description, denoted as $s \in \mathcal{R}^{1 \times 128}$.

\textbf{3D object encoder.}
We use the combined 3D data consisting of point coordinates and colors to represent the point cloud of each object on the 3D map.
To extract object features from the point cloud of the object candidates, we encode them using SparseConv~\cite{appendix_Graham_2018_CVPR}. We use average pooling to obtain object features denoted as $O \in \mathcal{R}^{n_o \times 128}$, where $n_o$ represents the number of object candidates targeted for visual grounding.
Similarly, we encode landmark objects using SparseConv, resulting in landmark features $L_o \in \mathcal{R}^{n_l \times 128}$. Here, $n_l$ denotes the number of landmarks present in the 3D map. Furthermore, we tokenize and encode the landmark names (e.g., `Clare College Conferencing') using a BiGRU. We obtain the sentence embeddings of the landmark names as landmark name features, denoted as $L_n \in \mathcal{R}^{n_l \times 128}$.
To incorporate both landmark object and name features, we concatenate and fuse them using a multi-layer perceptron. This fusion process yields the final landmark features $L \in \mathcal{R}^{n_l \times 128}$.

\textbf{Object localization module.}
The fused features, combining the object and landmark features, are further combined with the sentence embeddings of the input descriptions. The resulting fused features are then fed into a BiGRU to establish associations between the object and landmark features. Subsequently, a softmax function is applied to the output of the BiGRU within the object localization module. This step generates scores for the object candidates.
As a result, the CityRefer model outputs the object candidate with the highest score as the target corresponding to the given description.

\subsection{Training}
To train our 3D visual grounding model, we use a dataset consisting of 35k natural language descriptions of 3D objects. During training, we use the Adam optimizer~\cite{appendix_kingma2014adam} with a learning rate of 1e-4. 
The training process is executed on a single V100 GPU, using FP16 mixed precision through the PyTorch native amp module.
All hyperparameters are summarized in Table~\ref{table:3dvg_hyperparameter}.

\begin{table*}[h]
 \caption{Hyperparameters for training 3D visual grounding model.}
 \label{table:3dvg_hyperparameter}
 \centering
  \begin{tabular}{r|l}
   \toprule
    Hyperparameter & Value \\
    \midrule
    Training epochs & 30 \\
    Optimizer & Adam~\cite{appendix_kingma2014adam} \\
    Learning rate & 1e-4 \\
    Weight decay & 0.5 \\
    Batch size per GPU & 64 \\
    Number of object candidates & 10 \\
    Number of instance points & 1024 \\
    Hidden size & 128 \\
    Dropout probability & 0.1 \\
    Tokenizer & BertTokenizer \\
    Point cloud encoder & Sparse Convolutional Networks~\cite{appendix_Graham_2018_CVPR} \\
   \bottomrule
  \end{tabular}
\end{table*}

\begin{table*}[t]
\centering
\caption{Ablation study of the proposed baseline method with different features.}
\small
\begin{tabular}{l|cccc|c}
\toprule
Method  & Building &  Car & Ground & Parking & Overall \\
\midrule
Ours (point=512) & {0.252} $\pm$ 0.002 & {0.288} $\pm$ 0.006 & {0.469} $\pm$ 0.018 & \textbf{0.879} $\pm$ 0.014 & {0.314} $\pm$ 0.002\\
Ours (point=2048)  & {0.229} $\pm$ 0.003 & {0.291} $\pm$ 0.006 & {0.473} $\pm$ 0.016 & {0.846} $\pm$ 0.023 & {0.302} $\pm$ 0.004\\
Ours (wo/ color)  & {0.235} $\pm$ 0.006 & {0.287} $\pm$ 0.007 & {0.466} $\pm$ 0.013 & {0.845} $\pm$ 0.033 & {0.303} $\pm$ 0.004 \\
Ours (wo/ name)  & {0.250} $\pm$ 0.007 & {0.283} $\pm$ 0.009 & {0.475} $\pm$ 0.005 & {0.836} $\pm$ 0.020 & {0.310} $\pm$ 0.005 \\
\midrule
Ours  & \textbf{0.255} $\pm$ 0.008 & \textbf{0.298} $\pm$ 0.007 & \textbf{0.489} $\pm$ 0.009 & {0.853} $\pm$ 0.020 & \textbf{0.320} $\pm$ 0.005\\
\bottomrule
\end{tabular}
\label{tab:add_results_3dvg}
\end{table*}

\section{Additional Quantitative Analysis}
We describe ablation studies conducted on the CityRefer model.
Table~\ref{tab:add_results_3dvg} shows the results of the ablation study using our baseline method (Baseline + Land) with different features.

\noindent \textbf{Effect of instance size:}
We compared our baseline method (Ours), which uses 1024 points, with variants trained using 512 and 2048 points (Ours point=512, 2048).
The results indicate that the choice of the number of points in an instance affects the performance of the 3D visual grounding model.

\noindent \textbf{Effect of point colors:} 
In our evaluation, we compared the performance of our baseline method (Ours) with a variant trained without RGB values (Ours wo/ color). The results, as shown in the table, clearly demonstrate the effectiveness of color information in city-level 3D visual grounding. The use of RGB values is beneficial, particularly when distinguishing similar objects, such as cars or buildings, where the color of the roof plays a significant role in differentiation.

\noindent \textbf{Effect of landmark name:}
We conducted a comparison between our baseline method (Ours) and a variant trained without landmark names (Ours wo/ name). The results reveal the crucial role played by landmark names in enhancing the accuracy of 3D visual grounding.

\section{Additional Qualitative Analysis}
\begin{figure*}
    \centering
    \includegraphics[width=1.0\linewidth]{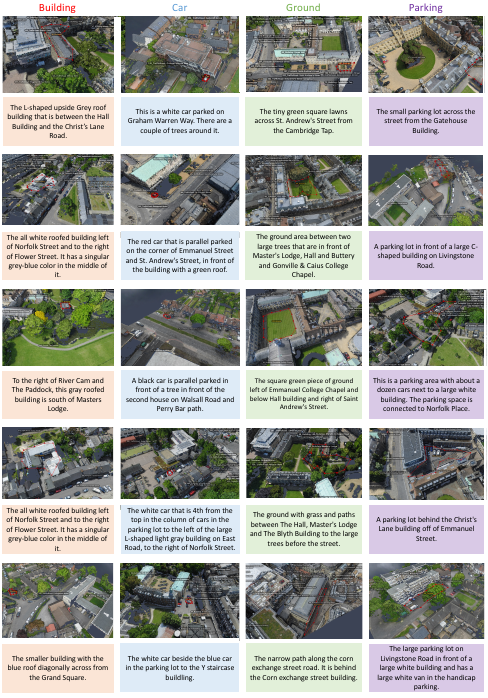}
    \caption{Additional qualitative examples. }
    \label{fig:additional_qualitative}
\end{figure*}

We demonstrate how our 3D visual grounding model works by visualizing examples. 
Figure~\ref{fig:additional_qualitative} shows several typical examples.
The results highlight the accurate prediction of the target object based on the provided descriptions, showing the discriminative ability of our model in the context of city-scale 3D visual grounding, thanks to the use of landmark features.
For example, in the second column of the first row, even in the presence of multiple white cars within the 3D data, our model effectively can use the geographic information of the road, `Graham Warren Way,' to narrow down the location of the target object while
a method without landmark information fails to make the correct prediction.

\clearpage

\end{document}